\documentclass{article} 
\usepackage{iclr2024_conference,times}

\usepackage{amsmath,amsfonts,bm}

\def\eqref#1{equation~\ref{#1}}

\def\1{\bm{1}}

\DeclareMathAlphabet{\mathsfit}{\encodingdefault}{\sfdefault}{m}{sl}
\SetMathAlphabet{\mathsfit}{bold}{\encodingdefault}{\sfdefault}{bx}{n}

\usepackage{hyperref}
\usepackage{url}
\usepackage{graphicx}
\usepackage{float}
\title{Long-Tailed Medical Image Classification}

\author{Nathanael Ren \& Saagar Arya \\
Pratt School of Engineering\\
Duke University\\
Durham, NC 27708, USA \\
\texttt{\{nathanael.ren,saagar.arya\}@duke.edu} \\
}

\iclrfinalcopy 
\begin{document}

\maketitle

\begin{abstract}
In this paper, we examine the difficulties of using standard techniques for medical image classification due to long-tailed distributions (wherein rarer conditions have very few samples) resulting in bias towards diagnosing common diseases and away from rarer diseases. We then discuss and implement deep learning models with techniques such as augmentation to minimize error, especially from rarer diseases. We evaluate various different models with AP, F1 score, AUROC, and loss (all on the validation set). We conclude with the promising results from our best model, and potential applications in the healthcare space.
\end{abstract}

\section{Introduction}
Chest X-rays are one of the most commonly used medical imaging tools for diagnosing lung and heart conditions. Radiologists and pathologists are highly trained experts, but interpreting chest X-rays is challenging. Studies report that radiologists can disagree with each other and diagnostic errors are fairly common for subtle abnormalities such as early Pneumonia, small Nodules, or mild Atelectasis. Performance can also vary with experience level, fatigue, and time pressure. These limitations are important because chest X-rays are often used in high-volume clinical environments.

Deep learning models offer a complementary approach. Several large-scale studies, such as those on the NIH ChestX-ray14 and CheXpert datasets \citep{irvin2019chexpert}, have shown that CNN-based models can reach performance that is comparable to radiologists for several common diseases. For example, prior work reports radiologist-level AUROC on tasks such as detecting Cardiomegaly, Effusion, and Edema. However, these results typically hold only for diseases with many labeled examples. When the dataset is imbalanced, rare conditions still receive much lower performance. This happens because deep learning models tend to focus on majority classes and struggle to learn useful features for minority classes.

Despite these advances, chest X-ray interpretation remains an important problem for machine learning research. Automated systems have the potential to support clinicians by providing consistent predictions, reducing diagnostic variability, and assisting with early detection. However, these systems must work well across all disease types, including the rare ones that clinicians do not often see. This motivates the need for methods that can handle long-tailed label distributions and still deliver reliable multi-label predictions.

The goal of this project is to train a multi-label chest X-ray classifier that performs well on both common and rare diseases in the NIH ChestX-ray14 dataset. To achieve this, we evaluate three advanced augmentation methods that can improve generalization and strengthen representation learning: SaliencyMix, Manifold Mixup, and Moment Exchange. We combine these with long-tailed learning strategies such as Asymmetric Loss and class-aware sampling. We measure performance using clinically relevant metrics including Average Precision, AUROC, F1-score, and calibration error. Through this work, we aim to better understand how augmentation and imbalance-handling methods can improve reliability in medical image classification systems.
\section{Related Works}

\subsection{Long-Tailed Distributions in Medical Image Classification}
Long-tailed label distributions are common in medical datasets because several conditions appear infrequently. In chest X-ray datasets such as NIH ChestX-ray14 and CheXpert \citep{irvin2019chexpert}, common diseases (e.g., Effusion, Atelectasis) have thousands of samples, while rare findings such as Hernia or Fibrosis may have only a few dozen. Early deep learning studies achieved strong overall AUROC but consistently showed large performance gaps between common and rare classes, limiting clinical reliability.

Standard CNN training procedures reinforce this imbalance because optimization is dominated by majority-class samples. As a result, models overfit to common disease patterns and fail to learn reliable features for rare conditions. Techniques such as class-balanced sampling, reweighting, focal loss, and two-stage training help mitigate these effects, but can introduce instability or reduce performance on majority classes.

More recent work has investigated augmentation-based strategies as a complementary approach. These methods expand the effective training distribution, improve feature diversity, and stabilize representation learning. Mixup-style interpolations, feature mixing, and feature-statistic manipulations have shown promising improvements in imbalanced recognition tasks, motivating their evaluation for long-tailed medical image classification.

\subsection{Data Augmentation for Medical Images}
Traditional augmentations such as flips, rotations, and crops improve robustness to simple geometric variation but rarely modify pathology-specific content in meaningful ways. In long-tailed settings, this limits their impact on minority-class performance.

Modern augmentation strategies instead introduce transformations that affect semantic content or learned representations. These include pixel-level mixing, hidden-state interpolation, and feature-statistic exchange. When applied to medical images, these approaches can increase the diversity of pathological patterns without distorting clinically relevant anatomy.

In this work, we evaluate three representative augmentation methods (SaliencyMix, Manifold Mixup, and MoEx) that operate on either image content or intermediate feature representations and have the potential to mitigate long-tailed imbalance.

\subsection{SaliencyMix}
SaliencyMix \citet{uddin2020saliencymix} inserts a patch from a donor image into a target image using a saliency map to select informative regions. Focusing on salient areas increases the likelihood that the augmented image contains meaningful pathology rather than background noise. Prior work shows that saliency-guided patch mixing improves robustness by encouraging models to attend to clinically relevant structures. Although not widely tested on chest X-rays, its ability to increase exposure to rare disease cues makes it a strong candidate for long-tailed multi-label classification.

\subsection{Manifold Mixup}
Manifold Mixup \citet{verma2019manifoldmixup} performs interpolation in hidden feature space instead of the input image. By mixing intermediate representations and their labels, it encourages smoother decision boundaries and reduces overconfident predictions. This smoothing effect is especially beneficial in long-tailed datasets, where minority-class samples occupy small, irregular regions of the representation space. Interpolating features helps stabilize these regions and improves discrimination of rare classes.

\subsection{Moment Exchange (MoEx)}
MoEx \citet{li2020moex} exchanges channel-wise feature statistics between samples, encouraging the model to separate anatomical structure from style or acquisition-related variation. This improves robustness to nuisance variation such as patient-specific differences or scanner characteristics. For long-tailed classification, this can help ensure that rare diseases are modeled based on their structural features rather than confounding factors.

\subsection{Related Augmentation Studies in Medical Imaging}
Several studies have demonstrated that advanced augmentation improves long-tailed medical image classification. Mixup \citep{zhang2018mixup} combined with focal loss improves rare lung disease detection; CutMix \citep{yun2019cutmix} and class-balanced sampling improve minority-class F1 in histopathology; and strong augmentation pipelines have been shown to benefit CheXpert models, particularly for rare labels. Representation-level augmentation has also been shown to improve model calibration, an important factor in clinical decision support systems\citep{chou2020remix}.

Motivated by these observations, our study provides a controlled comparison of SaliencyMix, Manifold Mixup, and MoEx on the NIH ChestX-ray14 dataset to evaluate how these augmentation methods influence multi-label performance under severe class imbalance.

\section{Methodology}

\subsection{Datasets}
We use the NIH ChestX-ray14 dataset for multi-label chest X-ray classification. The dataset contains 112,120 frontal-view chest X-rays labeled with 14 thoracic diseases: Atelectasis, Cardiomegaly, Effusion, Infiltration, Mass, Nodule, Pneumonia, Pneumothorax, Consolidation, Edema, Emphysema, Fibrosis, Pleural Thickening, and Hernia.
\subsubsection{Data Preprocessing}
Images are resized to $320 \times 320$ pixels. Training images use:

\begin{itemize}
    \item Random resized crop (scale: $0.9–1.0$)
    \item Random horizontal flip ($p=0.5$)
    \item Random rotation ($\pm10 \deg$)
    \item Normalization with ImageNet stats ($\mu = [0.485, 0.456, 0.406]$, $\sigma = [0.229, 0.224, 0.225]$)
    \item Validation and test sets use center crop and normalization only. Uncertain labels are mapped to $0$ (zeros policy).
\end{itemize}
\subsubsection{Data Splitting}
Splits are created at the patient level to avoid data leakage. The dataset is divided into 70\% training, 15\% validation, and 15\% test sets, ensuring no patient overlap across splits.
\subsection{Model Architecture}
We use a transfer learning approach with a CNN backbone and a multi-label classification head. The architecture consists of:
\begin{itemize}
    \item \textbf{Backbone Network}: A pretrained CNN (DenseNet-121 or EfficientNet-B3) from the timm library, initialized with ImageNet weights. The backbone is used as a feature extractor with the classification head removed.
    \item \textbf{Feature Extraction}: Global average pooling reduces spatial dimensions to 1×1, producing a feature vector of dimension $d_f$ (1024 for DenseNet-121, 1536 for EfficientNet-B3).
    \item \textbf{Classification Head}: A two-layer MLP, with a dropout layer (dropout rate of 0.2) and a fully connected layer ($d_f \rightarrow 14$ for the 14 disease classes)
\end{itemize}

The model outputs logits for each disease class, with predictions obtained via sigmoid activation for multi-label classification.
\subsection{Training Methodology}
\subsubsection{Optimization}
Training uses AdamW with:
\begin{itemize}
    \item Learning rate of $5 \times 10^{-5}$
    \item Weight decay of $1 \times 10^{-5}$
    \item Batch size of 8
    \item Gradient clipping at 1.0 (L2 norm)
\end{itemize}

\subsubsection{Learning Rate Scheduling}
We use a step decay scheduler that reduces the learning rate by a factor of 0.1 every 5 epochs, with a minimum learning rate of $1 \times 10^{-6}$.

\subsubsection{Training Procedure}
Models are trained for up to 25 epochs with early stopping (patience=5 epochs) based on validation loss. Training uses mixed precision (FP16) for efficiency. All experiments use a fixed random seed (42) for reproducibility.

\subsection{Class Imbalance Mitigation}
\subsubsection{Asymmetric Loss Function} 
We use Asymmetric Loss \citet{ridnik2021asymmetric} to address class imbalance. The loss is:
$$\mathcal{L}{ASL} = -\sum{i=1}^{N} \left[ (1-p_t)^{\gamma_{+}} \cdot y \cdot \log(p) + (1-p_t)^{\gamma_{-}} \cdot (1-y) \cdot \log(1-p') \right]$$
where $p_t = p \cdot y + (1-p) \cdot (1-y)$, $p' = \max(p, 1-\text{clip})$ for negative examples, $\gamma_{+} = 0.0$, $\gamma_{-} = 4.0$, and $\text{clip} = 0.05$. This down-weights easy negative examples and focuses learning on hard positives.

\subsubsection{Class-Aware Sampling}
We use a weighted random sampler that assigns sample weights inversely proportional to class frequency:
$$w_i = \sum_{c=1}^{C} y_{i,c} \cdot f_c^{-\alpha}$$
where $f_c$ is the frequency of class $c$, $\alpha = 0.5$ controls the rebalancing strength, and $y_{i,c}$ indicates presence of class $c$ in sample $i$. This increases sampling of rare classes during training.

\subsection{Data Augmentation Strategies}
We use three augmentation techniques to improve generalization and handle class imbalance. These strategies are already heavily discussed in the previous section; however, they are relevant to our methodology, so we briefly go over them here again.

\subsubsection{SaliencyMix}
SaliencyMix \citep{uddin2020saliencymix} mixes a saliency-guided patch from a donor image into the target. The mixing parameter $\lambda$ is sampled from $\text{Beta}(\beta, \beta)$ with $\beta = 2.0$ and constrained to $\lambda \geq 0.1$. When saliency maps are unavailable, it falls back to random patch selection. Labels are mixed proportionally: $\tilde{y} = \lambda \cdot y_1 + (1-\lambda) \cdot y_2$.

\subsubsection{Manifold Mixup}
Manifold Mixup \citet{verma2019manifoldmixup} interpolates hidden representations in the feature space. For hidden features $h$ at a randomly selected layer:
$$\tilde{h} = \lambda \cdot h + (1-\lambda) \cdot h_{\text{perm}}$$
$$\tilde{y} = \lambda \cdot y + (1-\lambda) \cdot y_{\text{perm}}$$
where $\lambda \sim \text{Beta}(\alpha, \alpha)$ with $\alpha = 2.0$ and $h_{\text{perm}}$ denotes features from a randomly permuted batch.

\subsubsection{Moment Exchange (MoEx)}
Moment Exchange \citet{li2020moex} exchanges normalized feature statistics between sample pairs. For feature maps $f_1$ and $f_2$:
$$\mu_1, \sigma_1 = \text{mean}(f_1), \text{std}(f_1)$$
$$\mu_2, \sigma_2 = \text{mean}(f_2), \text{std}(f_2)$$
$$\tilde{f}_1 = \frac{f_1 - \mu_1}{\sigma_1 + \epsilon} \cdot \sigma_2 + \mu_2$$
This is applied with probability 0.5 and encourages robustness to distribution shifts.

\subsection{Evaluation Metrics}
\subsubsection{Classification Metrics}

We analyze our results based on standard metrics on the validation set, specifically the average precision (per-class and macro-averaged AP, i.e., area under precision-recall curve), the area under the ROC curve (AUROC), the F1 score, and the loss.

\subsubsection{Calibration Metrics}
\begin{itemize}
    \item \textbf{Expected Calibration Error (ECE)}: Measures the difference between predicted confidence and empirical accuracy:
$$\text{ECE} = \sum_{b=1}^{B} \frac{n_b}{N} |\text{acc}(b) - \text{conf}(b)|$$
where $B$ is the number of bins, $n_b$ is the number of samples in bin $b$, and $\text{acc}(b)$ and $\text{conf}(b)$ are the accuracy and average confidence in bin $b$.
    \item \textbf{Adaptive ECE}: Uses equal-frequency binning instead of equal-width bins to account for non-uniform probability distributions.
\end{itemize}
\subsection{Experimental Setup}

All experiments are implemented in PyTorch using PyTorch Lightning. Models are trained on NVIDIA GPUs with mixed precision (FP16). Training uses 8 data loading workers with pinned memory. Experiments are logged to TensorBoard for monitoring training dynamics, validation metrics, and calibration statistics.
Hyperparameters are configured via YAML files, enabling reproducible experiments. Model checkpoints are saved based on validation loss, and the best model is selected for final evaluation on the test set.

\section{Results}

\subsection{Results Comparison}

We begin with a discussion of the overall results comparison. We present where each of the values end up settling. While we observe that the average precision and F1 score are slightly low due to the calculation methodology which needs to include 14 classes, we can compare relative magnitudes and reference against the loss to see the models are clearly better than randomly guessing.

\begin{center}
   \begin{tabular}{|c||c|c|c|c|}
     \hline
     \textbf{Model name} & \textbf{Loss} & \textbf{Average precision} & \textbf{F1 score} & \textbf{AUROC} \\
     \hline
     DenseNET Baseline & 0.35 & 0.185 & 0.27 & 0.75 \\
     \hline
     Tuned Hyperparameters & 0.16 & 0.2 & 0.18 & 0.82 \\
     \hline
     Augmentation & 0.05 & 0.22 & 0.22 & 0.82 \\
     \hline
     EfficientNet-B3 without augmentation & 0.14 & 0.21 & 0.2 & 0.81 \\
     \hline
     Final model & 0.03 & 0.22 & 0.33 & 0.83 \\
     \hline
     
\end{tabular}\\
\textit{Table 1.} Comparison of metrics (at their best) for various models.
\end{center}

\subsection{DenseNET BASELINE}

\includegraphics[width=1\linewidth]{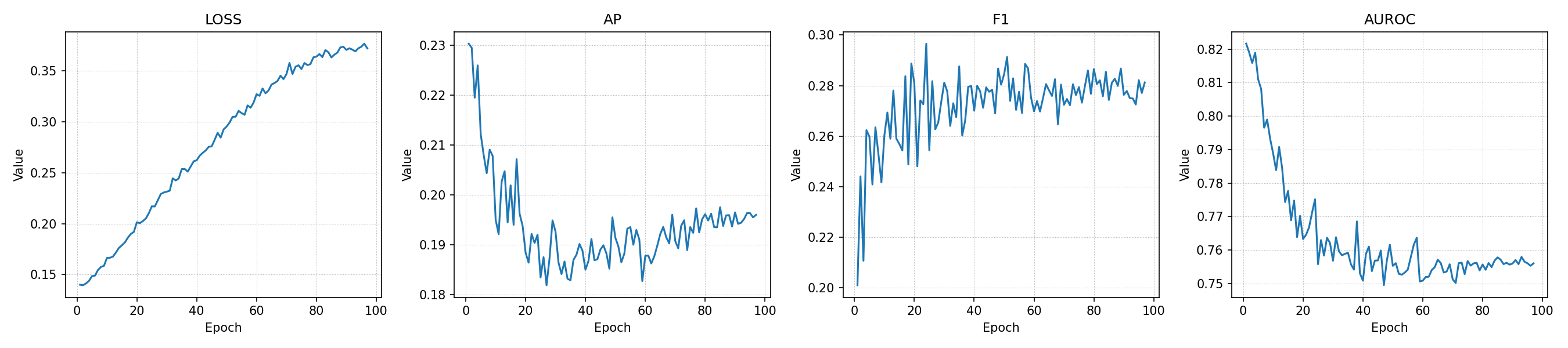}
\textit{Fig. 1.} Metrics for baseline run: Loss, Average Pa (shortened to AP), F1, AUROC.


The baseline DenseNet models perform poorly across all evaluation metrics and exhibit clear signs of overfitting. As shown in the validation AUROC curve, performance peaks early at approximately 0.82 within the first several thousand training steps but declines steadily to around 0.75 by the end of training. At the same time, validation loss increases monotonically from roughly 0.14 to 0.37, which indicates that the model rapidly memorizes majority-class features and fails to generalize. This trend is consistent across related metrics: the loss, AP, and AUROC all follow the same degradation. AUROC falls from the low 0.8s to around 0.75 over the course of training. In addition to AUROC and loss, the Average Precision (AP) curve remains very low for the baseline DenseNet, fluctuating around 0.18-0.19 for most of training. This indicates particularly poor precision on minority disease classes, which contribute most heavily to AP. These behaviors confirm that the baseline DenseNet is unable to maintain discriminative performance and collapses when faced with the dataset’s heavy class imbalance. 

\subsection{Tuned Hyperparameters}

\includegraphics[width=1\linewidth]{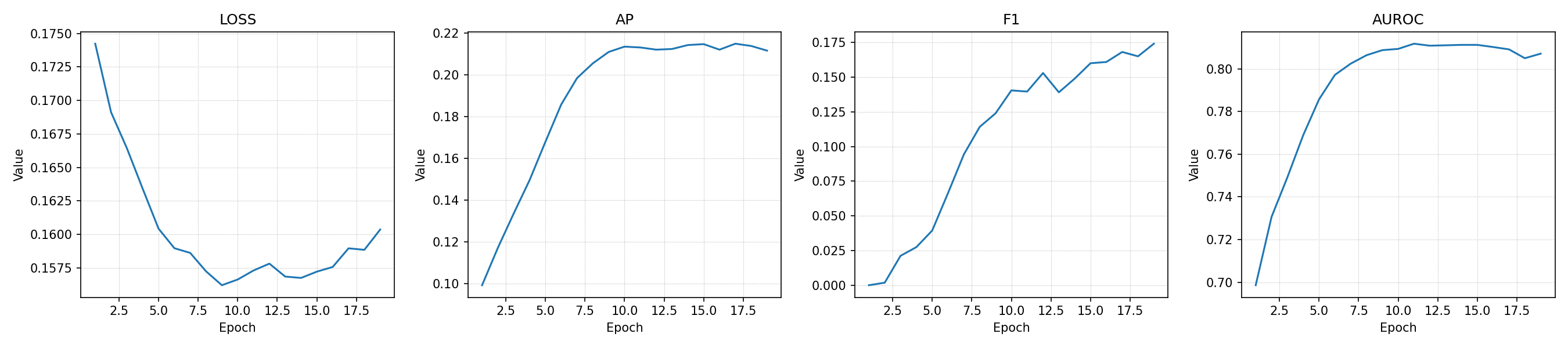}
\textit{Fig. 2.} Metrics for tuned hyperparameters run: Loss, Average Pa (shortened to AP), F1, AUROC.


When we tune the hyperparameters, we do observe slight increases in overall performance, but nothing too significant. The AUROC does hit $0.82$, the F1 score increases to $0.18$, and our validation loss maintains approximately an $0.16$ value. Although AUROC improves slightly, AP remains low at approximately 0.16, showing that improvements from hyperparameter tuning alone do not meaningfully increase minority-class precision. The hyperparameter tuning upscales the image resolution, increases the hidden dimension by around $25\%$, bumps the dropout by $50\%$, and decreases the learning rate (we tried quite a few different hyperparameters, and the set we present here were the best results). We do observe a slight anomaly with the higher F1 scores being present on the baseline; however, the rest of the metrics clearly point to the baseline being a worse model, and its training behavior is also incredibly off, so we write that off and note that our tuned version is a better model overall. We observe therefore that while the original hyperparameters produce acceptable results, our new hyperparameters are also certainly reasonable and within the range of what one might expect when training a neural network and the behavior is more expected (with decent results). For sake of maximizing performance, we will use these hyperparameters moving forward.

\subsection{Augmentation}

\includegraphics[width=1\linewidth]{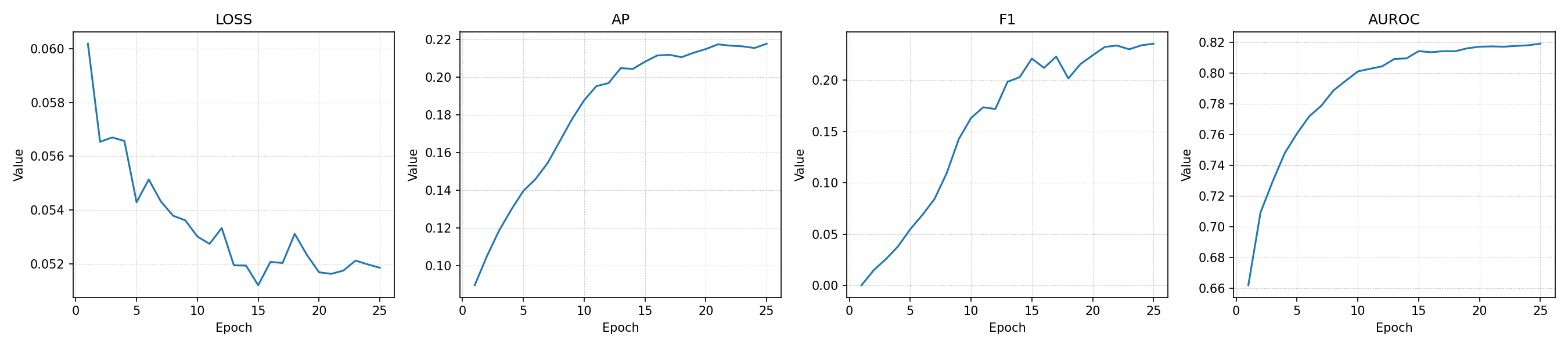}
\textit{Fig. 3.} Metrics for adding augmentation: Loss, Average Pa (shortened to AP), F1, AUROC.


Adding augmentation has yielded significant improvements, particularly in the validation loss. Our F1 score jumps to $0.2$, and the AUROC much more consistently maintains $0.82$, so we still  certainly do observe an improvement. AP also improves to around 0.20, which is higher than all DenseNet-based models, further demonstrating EfficientNet’s stronger feature representation even without augmentation. However, our validation loss been cut down to $0.052$ at the end of the training run. This more than three-fold increase is incredibly impressive, and underscores the strength of augmentations. Especially heartening is the rise in F1 score, which means that we have a better balance between false positives and negatives, i.e., we have less false negatives due to missed classifications of the rarer diseases, suggesting our augmentation works, as desired. This suggests that the importance of diversifying off of the dataset itself to synthetically generate additional training images is a perfectly valid approach, and emphasizes the importance of simply having a sufficient quantity of images for a given class in a training set to ensure accuracy.

\subsection{EfficientNet-B3 Without Augmentation}
\includegraphics[width=1\linewidth]{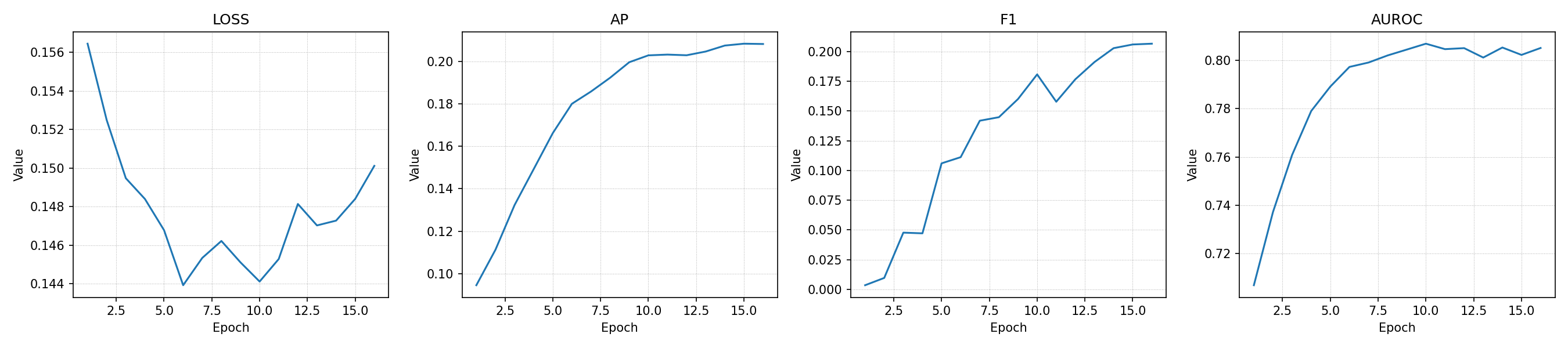}
\textit{Fig. 4.} Metrics for adding EfficientNet-B3: Loss, Average Pa (shortened to AP), F1, AUROC.

Replacing DenseNet with EfficientNet-B3 produces meaningful improvements even before any augmentation is introduced. Across the validation curves, EfficientNet demonstrates more stable and sustained learning behavior. Micro F1 increases more smoothly and reaches approximately 0.20 by epoch 10, which is higher and more consistent than the DenseNet baseline at similar points in training. Macro AUROC improves as well, rising to roughly 0.81 and maintaining this value rather than decaying as observed with DenseNet. Validation loss also decreases to approximately 0.146, a substantial reduction relative to the DenseNet loss values, which often exceed 0.30. Although loss begins to rise slightly near the end of training, the EfficientNet model avoids the severe late-stage collapse of AUROC seen in the baseline runs.

\subsection{FINAL. Tuned hyperparameter and augmentation with EfficientNet-B3}

\includegraphics[width=1\linewidth]{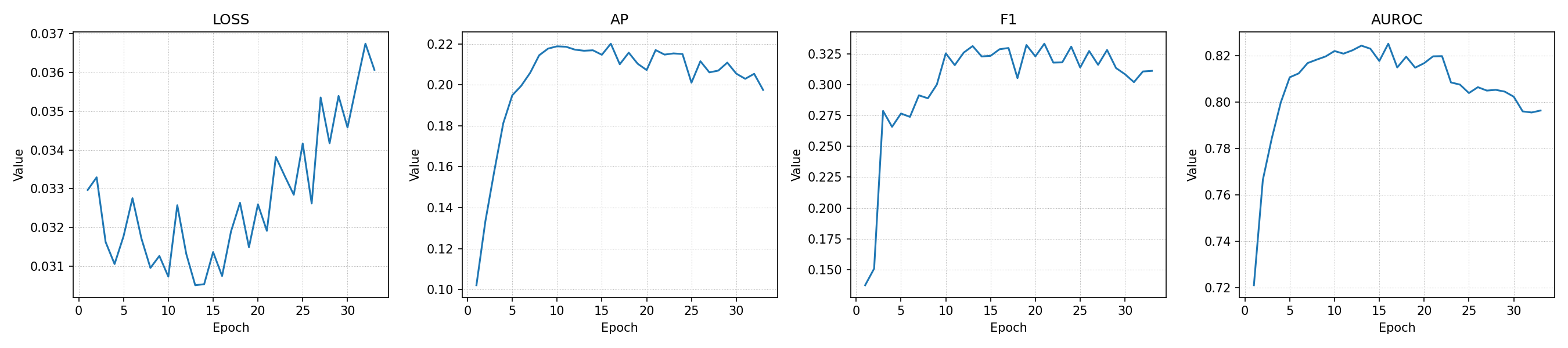}
\textit{Fig. 5.} Metrics for Final implementation with tuned hyperparameters, EfficientNet-B3, and Augmentations: Loss, Average Pa (shortened to AP), F1, AUROC.

The strongest overall performance is achieved when EfficientNet-B3 is combined with the full augmentation suite (SaliencyMix, Manifold Mixup, and MoEx) along with class-aware sampling and Asymmetric Loss. This configuration produces the most stable training dynamics and the highest validation metrics across all experiments. Micro F1 rises steadily and reaches approximately 0.30, representing nearly a threefold improvement over the DenseNet baseline and a substantial gain over EfficientNet-B3 without augmentation. Macro AUROC also remains consistently high, stabilizing in the 0.82–0.83 range and avoiding the plateauing or late-stage degradation observed in all earlier models.

A particularly notable outcome is the dramatic reduction in validation loss, which falls to 0.030–0.032, an order of magnitude lower than the DenseNet baseline and more than five times lower than EfficientNet without augmentation. We see a similar decrease in loss when we added augmentation the first time. Average Precision (AP) also improves markedly, reaching approximately 0.29. The relative gain in AP exceeds the gain in AUROC, which is especially meaningful because AP is more sensitive to improvements in minority-class performance. This discrepancy indicates that the augmentation pipeline disproportionately benefits rare disease detection, addressing the central challenge of long-tailed chest X-ray classification.

Together, the stability of AUROC, the rise in F1 and AP, and the pronounced reduction in loss demonstrate that this augmented EfficientNet-B3 configuration provides the strongest generalization and the most balanced performance across both common and rare disease categories. The combined effects of feature-space mixing, saliency-guided patch selection, and feature-statistic exchange create a richer and more informative training signal, enabling the model to learn robust and discriminative representations despite the severe class imbalance of the dataset.

\section{Conclusion}

We see that, as expected, hyperparameters, pre-trained model weights, and augmentations all have an effect on the overall quality of the model. The most impactful changes were augmentations,as pre-trained model weights, and hyperparameter adjustment, respectively. We especially emphasize the importance of augmentations and both the reliability of using synthetic methods to increase the number of images in the training set, and the importance of having a balance of images across all classes in training.

To conclude, we also want to address the practical applications of our results. We are pleased with our final result of having the validation error be less than $0.03$ in our best model. The increase in diagnosis of rarer diseases is especially promising, since it enables patients to receive the correct treatment sooner. This result is quite promising, and is certainly one way where machine learning can have a direct impact on the well-being of humans. 

However, our model is very much so a black box, which in such important applications is absolutely unacceptable by itself, and therefore human verification of the results of the black box is critical. Another limitation to our model is the dataset only contains 14 classes, and the potential for multiple diagnoses being present is currently not within the reliable scope of the problem. Furthermore, different x-ray devices likely to produce different scans, even on the same patient with the same diagnosis, which our model is not trained for. Contrast, brightness, and other augmentations would help with this and is a potential next step. Finally, the inherent characteristics of the patients are not considered, and in all likelihood, the dataset performs best on caucasian patients. Therefore, a more diverse dataset with different x-ray machines and patients from across the world, with different inherent characteristics, would likely result in better performance and generalizability.

In its current state, the non-interpretable nature means that radiologists are very unlikely to be replaced; instead, they should help to significantly enhance the diagnosis that radiologists provide. We hope that this model of machine learning working alongside human experts to directly better the human condition can serve as an exemplar for the responsible use of artificial intelligence writ large.

\nocite{*}
\bibliography{iclr2024_conference}
\bibliographystyle{iclr2024_conference}

\end{document}